# Failure Modes in Machine Learning

Ram Shankar Siva Kumar*, David O'Brien#, Kendra Albert^, Salome Viljoen#, Jeffrey Snover*

ram.shankar@microsoft.com; jsnover@microsoft.com        dobrien@cyber.harvard.edu; sviljoen@cyber.harvard.edu        kalbert@law.harvard.edu
*Microsoft        #Berkman Klein Center for Internet and Society at Harvard University        ^Harvard Law School

## Introduction & Background

In the last two years, more than 200 papers have been written on how machine learning (ML) can fail because of adversarial attacks on the algorithms and data; this number balloons if we were to incorporate papers covering non-adversarial failure modes. The spate of papers has made it difficult for ML practitioners, let alone engineers, lawyers, and policymakers, to keep up with the attacks against and defenses of ML systems. However, as these systems become more pervasive, the need to understand how they fail, whether by the hand of an adversary or due to the inherent design of a system, will only become more pressing. The purpose of this document is to jointly tabulate both of these failure modes in a single place.

- *Intentional failures* where the failure is caused by an active adversary attempting to subvert the system to attain her goals – either to misclassify the result, infer private training data, or to steal the underlying algorithm.
- *Unintentional failures* where the failure is because an ML system produces a formally correct but completely unsafe outcome

We would like to point out that there are other taxonomies and frameworks that individually highlight intentional failure modes[1,2] and unintentional failure modes.[3,4] Our classification brings the two separate failure modes together in one place and addresses the following needs:

1) The need to equip software developers, security incident responders, lawyers, and policy makers with a common vernacular to talk about this problem. After developing the initial version of the taxonomy last year, we worked with security and ML teams across Microsoft, 23 external partners, standards organization, and governments to understand how stakeholders would use our framework. Based on this usability study and stakeholder feedback, we iterated on the framework.
   Result: When presented with an ML failure mode, we frequently observed that software developers and lawyers mentally mapped the ML failure modes to traditional software attacks like data exfiltration. So, throughout the paper, we attempt to highlight how machine learning failure modes are meaningfully different from traditional software failures from a technology and policy perspective.
2) The need for a common platform for engineers to build on top of and to integrate into their existing software development and security practices. Broadly, we wanted the taxonomy to be more than an educational tool – we want it to effectuate tangible engineering outcomes.

---

[1] Li, Guofu, et al. "Security Matters: A Survey on Adversarial Machine Learning." *arXiv preprint arXiv:1810.07339* (2018).
[2] Chakraborty, Anirban, et al. "Adversarial attacks and defences: A survey." *arXiv preprint arXiv:1810.00069* (2018).
[3] Ortega, Pedro, and Vishal Maini. "Building safe artificial intelligence: specification, robustness, and assurance." *DeepMind Safety Research Blog* (2018).
[4] Amodei, Dario, et al. "Concrete problems in AI safety." *arXiv preprint arXiv:1606.06565* (2016).

Result: Using this taxonomy as a lens, Microsoft modified its software development lifecycle process for its entire organization. Specifically, all engineers in Microsoft building ML systems now have a [threat model](#) to ascertain threats to their ML systems before deploying it in production and also security incident responders have a standard process for classifying ML vulnerabilities that have been discovered

3) The need for a common vocabulary to describe these attacks amongst policymakers and lawyers. We believe that this for describing different ML failure modes and analysis of how their harms might be regulated is a meaningful first step towards informed policy.

    Result: This taxonomy is written for a wide interdisciplinary audience – so, policymakers who are looking at the issues from a general ML/AI perspective, as well as specific domains such as misinformation/healthcare should find the failure mode catalogue useful. We also highlight any applicable legal interventions to address the failure modes.

## How to use this document?

At the outset, we acknowledge that this is a ***living document***. Since this is an active area of research, the document may not be complete; as the community discovers more failures modes, we will update the taxonomy. We also do not prescribe technological mitigations to these failure modes, as defenses are scenario specific and tie in with the threat model under consideration.

For engineers, we recommend browsing through the overview of possible failure modes and jumping into the [threat model](#) document. This way, engineers can identify threats, attacks and vulnerabilities and then use the framework to plan for countermeasures where available. We then refer you to the [bug bar](#) that maps these new vulnerabilities in the taxonomy alongside traditional software vulnerabilities, and provides a rating for each ML vulnerability (such as critical, important). This bug bar can be easily integrated with existing incident response process, triggering similar playbooks

For lawyers and policy makers, this document organizes ML failure modes into a framework to allow for analysis of key issues relevant for anyone exploring policy options, such as the work done here.[5,6] Specifically, we have categorized failures and consequences so that policy makers can begin to draw distinctions between causes which will in turn inform public policy initiatives to promote ML safety and security. We hope that policy makers will use these categories begin to flesh out how existing legal regimes may (not) adequately capture emerging issues, what historical legal regimes or policy solutions might have dealt with similar harms, and where we should be especially sensitive to civil liberties issues.

## Structure of the document

In both the *Intentional Failure Modes* and *Unintentional Failure Modes* section, we provide a brief definition of the attack, and illustrative example from the literature.

In the *Intentional Failure Modes* section, we provide the additional fields:

1) **What does the attack attempt to compromise in the ML system – Confidentiality, Integrity or Availability?** We define Confidentiality as assuring that the components of the ML system (data, algorithm, model) are accessible only by authorized parties; Integrity is defined as assuring that the ML system can be modified only by authorized parties; Availability is defined as an assurance that the ML system is accessible to authorized parties. Together, Confidentiality, Integrity and Availability is called the CIA triad. For each intentional failure mode, we attempt to identify which of the CIA triad is compromised.

---

[5] Shankar Siva Kumar, Ram, et al. "Law and Adversarial Machine Learning." *arXiv preprint arXiv:1810.10731* (2018).
[6] Calo, Ryan, et al. "Is Tricking a Robot Hacking?." University of Washington School of Law Research Paper 2018-05 (2018).

2) **How much knowledge is required to mount this attack – blackbox or whitebox?** In Blackbox style attacks., the attacker does NOT have direct access to the training data, no knowledge of the ML algorithm used and no access to the source code of the model. The attacker only queries the model and observes the response. In a whitebox style attack the attacker has knowledge of either ML algorithm or access to the model source code.

3) Commentary on if the attacker is violating traditional technological notion of access/authorization.

## Summary

| | Intentionally Motivated Failures | | |
|---|---|---|---|
| S.no | Attack | Overview | Violates traditional technological notion of access/authorization? |
| 1 | Perturbation attack | Attacker modifies the query to get appropriate response | No |
| 2 | Poisoning attack | Attacker contaminates the training phase of ML systems to get intended result | No |
| 3 | Model Inversion | Attacker recovers the secret features used in the model by through careful queries | No |
| 4 | Membership Inference | Attacker can infer if given data record was part of the model's training dataset or not | No |
| 5 | Model Stealing | Attacker is able to recover the model by constructing careful queries | No |
| 6 | Reprogramming ML system | Repurpose the ML system to perform an activity it was not programmed for | No |
| 7 | Adversarial Example in Physical Domain | Attacker brings adversarial examples into physical domain to subvert ML system e.g: 3d printing special eyewear to fool facial recognition system | No |
| 8 | Malicious ML provider recovering training data | Malicious ML provider can query the model used by customer and recover customer's training data | Yes |
| 9 | Attacking the ML supply chain | Attacker compromises the ML models as it is being downloaded for use | Yes |
| 10 | Backdoor ML | Malicious ML provider backdoors algorithm that doesn't work unless triggered | Yes |
| 11 | Exploit Software Dependencies | Attacker uses traditional software exploits like buffer overflow to confuse ML systems | Yes |

| | Unintended Failures | |
|---|---|---|
| S.no | Failure | Overview |
| 1 | Reward Hacking | Reinforcement Learning (RL) systems act in unintended ways because of mismatch between stated reward and true reward |
| 2 | Side Effects | RL system disrupts the environment as it tries to attain its goal |

| 3 | Distributional shifts | The system is tested in one kind of environment, but is unable to adapt to changes in other kinds of environment |
| 4 | Natural Adversarial Examples | Without attacker perturbations, the ML system fails owing to hard negative mining |
| 5 | Common Corruption | The system is not able to handle common corruptions and perturbations such as tilting, zooming, or noisy images. |
| 6 | Incomplete Testing | The ML system is not tested in realistic conditions that it is meant to operate in. |

| | Part I: Intentionally Motivated Failures | | | | |
|---|---|---|---|---|---|
| S.No | Attack Class | What | Compromises | Scenario | Is the attacker [technologically] misusing the system? |
| 1 | Perturbation attacks | In perturbation style attacks, the attacker stealthily modifies the query to get a desired response | ML-Integrity | Image: Noise is added to an X-ray image, which makes the predictions go from normal scan to abnormal. [7] **[Blackbox]**<br><br>Text translation: Specific characters are manipulated to result in incorrect translation. The attack can suppress specific word or can even remove the word completely.[8] **[Blackbox and Whitebox]**<br><br>Speech: Researchers showed how given a speech waveform, another waveform can be exactly replicated but transcribes into a totally different text.[9] [**Whitebox but may be extended to Blackbox**] | Technologically, it feels like the attacker is not misusing the system<br><br>- In a blackbox setting, no special privileges are required by the attacker to perform the attack. The attacker generates the perturbations offline, and queries the system legitimately.<br><br>- There seems to be no technological access violations – Just like a legitimate user sends in a legitimate image, the attacker sends in a corrupted image for classification to purposely confuse the system. |

---

[7] Paschali, Magdalini, et al. "Generalizability vs. Robustness: Adversarial Examples for Medical Imaging." arXiv preprint arXiv:1804.00504 (2018).

[8] Ebrahimi, Javid, Daniel Lowd, and Dejing Dou. "On Adversarial Examples for Character-Level Neural Machine Translation." arXiv preprint arXiv:1806.09030 (2018)

[9] Carlini, Nicholas, and David Wagner. "Audio adversarial examples: Targeted attacks on speech-to-text." arXiv preprint arXiv:1801.01944 (2018).

| 2 | Poisoning attacks | The goal of the attacker is to contaminate the machine model generated in the training phase, so that predictions on new data will be modified in the testing phase.<br><br>Targeted: In targeted poisoning attacks, the attacker wants to misclassify specific examples.<br><br>Indiscriminate: The aim here is to cause DoS like effect, which makes the system. | ML-Integrity | In a medical dataset where the goal is to predict the dosage of anticoagulant drug Warfarin using demographic information, etc. Researchers introduced malicious samples at 8% poisoning rate, which changed dosage by 75.06% for half of patients.[10] [**Blackbox**]<br><br>In Tay, future conversations were tainted because a fraction of the past conversations were used to train the system via feedback.[11] [**Blackbox**] | Technologically, it feels like the attacker is <u>not</u> misusing the system<br><br>- No special privileges required by the attacker. In a closed system like Twitter (where Tay, the tweetbot was corrupted), the attacker needs to be a user of the platform.<br><br>- No technological access violation – The authorized attacker sends chaff traffic to the endpoint, just like an authorized user would send legitimate traffic to the end point. |
| --- | --- | --- | --- | --- | --- |
| 3 | Model Inversion | The private features used in machine learning models can be recovered. | ML – Confidentiality | Researchers[12] were able to recover private training data used to train the algorithm. The authors were able to reconstruct faces, by just the name and access to the model to the point where workers on Amazon Mechanical Turk could use the photo to identify an individual from a line-up with 95% accuracy. | Technologically, it feels like the attacker is <u>not</u> misusing the system<br><br>- No special privileges are required by the attacker to perform the attack.<br><br>- There seems to be no technological access violations – Just like a legitimate user sends in a legitimate image, the |

---

[10] Jagielski, Matthew, et al. "Manipulating machine learning: Poisoning attacks and countermeasures for regression learning." *arXiv preprint arXiv:1804.00308* (2018)
[11] https://blogs.microsoft.com/blog/2016/03/25/learning-tays-introduction/
[12] Fredrikson M, Jha S, Ristenpart T. 2015. Model inversion attacks that exploit confidence information and basic countermeasures

| | | | | In the same paper, the authors showed how they were able to extract specific information. Verbatim from [12] | attacker sends these specially crafted queries to recover the private features. |
|---|---|---|---|---|---|
| | | | | *"In May 2014, Walt Hickey wrote an article for FiveThirtyEight's DataLab section that attempted a statistical analysis of the connection between peoples' steak preparation preferences and their propensity for risk-taking behaviors. To support the analysis, FiveThirtyEight commissioned a survey of 553 individuals from SurveyMonkey, which collected responses to questions such as: "Do you ever smoke cigarettes?", "Have you ever cheated on your significant other?", and of course, "How do you like your steak prepared?". Demographic characteristics such as age, gender, household income, education, and census region were also collected. We discarded rows that did not contain responses for the infidelity question or the steak preparation question, resulting in a total of 332 rows for the inversion experiments. <u>We used model inversion on the decision tree learned from this dataset to infer whether each participant responded "Yes" to the question about infidelity"</u>*<br><br>[**Whitebox** and **Blackbox**] | |
| 4 | Membership Inference attack | The attacker can determine whether a given data record was part of the model's training dataset or not. | ML – Confidentiality | Researchers were able to predict a patient's main procedure (e.g: Surgery the patient went through) based on the attributes (e.g: age, gender, hospital)[13] **[Blackbox]** | Technologically, it feels like the attacker is <u>not</u> misusing the system<br><br>- No special privileges are required by the attacker to perform the attack.<br><br>- There seems to be no technological access violations – Just like a legitimate user sends in a legitimate image, the attacker sends these specially crafted queries to infer membership. |
| 5 | Model stealing | The attackers recreate the underlying model by legitimately querying the model. The functionality of the new model is same as that of the underlying model. | ML-Confidentiality | Researchers successfully emulated the underlying algorithm from Amazon, BigML. For instance, in the BigML case, researchers were able to recover the model used to predict if someone should have a good/bad credit risk (German Credit Card dataset) using 1,150 queries and within 10 minutes[14] | Technologically, it feels like the attacker is <u>not</u> misusing the system<br><br>- No special privileges are required by the attacker to perform the attack.<br><br>- There seems to be no technological access violations – Just like a legitimate user sends in a legitimate image, the attacker sends these specially crafted queries to confuse the system |

---

[13] Shokri R, Stronati M, Song C, Shmatikov V. 2017. Membership inference attacks against machine learning models. In *Proc. of the 2017 IEEE Symp. on Security and Privacy (SP)*, *San Jose, CA, 22–24 May 2017*, pp. 3–18. New York, NY: IEEE.

[14] Tramèr, Florian, et al. "Stealing Machine Learning Models via Prediction APIs." *USENIX Security Symposium*. 2016.

| 6 | Reprogramming deep neural nets | By means of a specially crafted query from an adversary, machine learning systems can be reprogrammed to a task. that deviates from the creator's original intent. | ML – Integrity, ML-Availability | Demonstrated how ImageNet, a system used to classify one of several categories of images was repurposed to count squares. Authors end the paper with a hypothetical scenario: An attacker sends Captcha images to the computer vision classifier in a cloud hosted photos service to solve the image captchas to create spam accounts.[15] | Technologically, it feels like the attacker is <u>not</u> misusing the system<br><br>- No special privileges are required by the attacker to perform the attack.<br><br>- There seems to be no technological access violations – Just like a legitimate user sends in a legitimate image, the attacker sends these specially crafted queries to reprogram the system |
|---|---|---|---|---|---|
| 7 | Adversarial Example in the physical domain | An adversarial example is an input/query from a malicious entity sent with the sole aim of misleading the machine learning system. These examples can manifest in the physical domain. | ML - Integrity | Researchers 3D prints a rifle with custom texture that fools image recognition system into thinking it is a turtle[16]<br><br>Researchers construct a sunglass with a design that can now fool image recognition system, and no longer recognize the faces correctly[17] | See analysis above. |
| 8 | Malicious ML providers who can recover training data | Malicious ML provider can query the model used by customer and recover customer's training data. | ML – Confidentiality | Researchers show how a malicious provider presents a backdoored algorithm, wherein the private training data is recovered. They were able to reconstruct faces and texts, given the model alone. [18] | A trusted party (the provider) is misusing their access in order to cause harm.<br><br>However, depending on contractual agreements between a provider and user, there may or may not actually be an access or authorization problem. |

---

[15] Elsayed, Gamaleldin F., Ian Goodfellow, and Jascha Sohl-Dickstein. "Adversarial Reprogramming of Neural Networks." *arXiv preprint arXiv:1806.11146* (2018).

[16] Athalye, Anish, and Ilya Sutskever. "Synthesizing robust adversarial examples." *arXiv preprint arXiv:1707.07397*(2017)

[17] Sharif, Mahmood, et al. "Adversarial Generative Nets: Neural Network Attacks on State-of-the-Art Face Recognition." *arXiv preprint arXiv:1801.00349* (2017).

| 9 | Attacking the ML Supply Chain[19] | Owing to large resources (data + computation) required to train algorithms, the current practice is to reuse models trained by large corporations, and modify them slightly for task at hand (e.g: ResNet is a popular image recognition model from Microsoft). These models are curated in a Model Zoo (for example, Caffe hosts popular image recognition models). In this attack, the adversary attacks the models hosted in Caffe, thereby poisoning the well for anyone else. | ML - Integrity | Researchers show how it is possible for an attacker to check in malicious code into one of the popular model. An unsuspecting ML developer downloads this model and uses it as part of the image recognition system in their code. [20] The authors show how in Caffe, there exists a model whose SHA1 hash does NOT match the authors' digest, indicating tampering. There are 22 models without any SHA1 hash for integrity checks at all. | A trusted party (the attacker contributing code) is misusing their access in order to cause harm.<br><br>- Injecting malicious code into source repository feels like a technological violation. However, it may not map well onto existing legal regimes.<br><br>Questions:<br><br>1. Are model repositories (typically open source) failing their duty of care if they don't check for adversarial manipulation of the models they host?<br><br>2. Can the adversary who tampered with the models be punished under CFAA?<br><br>3. Is the unsuspecting developer now liable for any damages incurred by her/his/their customer? |
| 10 | Backdoor Machine Learning | Like in the "Attacking the ML Supply Chain", In this attack scenario, the training process is either fully or partially outsourced | ML – Confidentiality, ML-Integrity | Researchers created a backdoored U.S. street sign classifier that identifies stop signs as speed limits only when a special sticker is added to the stop sign (backdoor trigger). [20] | A trusted party is misusing their access in order to cause harm. However, it may not map well onto existing legal regimes and may depend on the contractual relationship between the malicious party |

---

[19] Xiao, Qixue, et al. "Security Risks in Deep Learning Implementations." *arXiv preprint arXiv:1711.11008* (2017).
[20] Gu, Tianyu, Brendan Dolan-Gavitt, and Siddharth Garg. "Badnets: Identifying vulnerabilities in the machine learning model supply chain." *arXiv preprint arXiv:1708.06733* (2017)

| | | | | | |
|---|---|---|---|---|---|
| | | to a malicious party who wants to provide the user with a trained model that contains a backdoor. The backdoored model would perform well on most inputs (including inputs that the end user may hold out as a validation set) but cause targeted misclassifications or degrade the accuracy of the model for inputs that satisfy some secret, attacker-chosen property, which we will refer to as the backdoor trigger. | | They are now extending this work to text processing systems, wherein specific words are replaced with the trigger being the speaker's accent.[21] | owning the training process and the people who commission the model. |
| 11 | Exploit software dependencies of ML system | In this attack, the attacker does NOT manipulate the algorithms. Instead, exploits the software vulnerabilities such as a buffer overflow within dependencies of the system. | ML- Confidentiality, ML – Integrity, ML- Availability | An adversary sends in corrupt input to an image recognition system that causes it to misclassify by exploiting a software bug in one of the dependencies.[3] | A trusted party is misusing their access in order to cause harm. |

---

[21] https://www.wired.com/story/machine-learning-backdoors/

## Part II: Unintended Failure Modes

| S.no | Attack Class | What | Compromises | Scenario |
|---|---|---|---|---|
| 1 | **Reward Hacking** | Reinforcement learning (RL) systems act in unintended ways because of discrepancies between the specified reward and the true intended reward. | Safety of the system | A huge corpus of gaming examples in AI has been compiled.[22] |
| 2 | **Side Effects** | RL system disrupts the environment as it tries to attain their goal | Safety of the system | Scenario, verbatim from the authors in [23]: *"Suppose a designer wants an RL agent (for example our cleaning robot) to achieve some goal, like moving a box from one side of a room to the other. Sometimes the most effective way to achieve the goal involves doing something unrelated and destructive to the rest of the environment, like knocking over a vase of water that is in its path. If the agent is given reward only for moving the box, it will probably knock over the vase."* |
| 3 | **Distributional shifts** | The system is tested in one kind of environment, but is unable to adapt to changes in other kinds of environment | Safety of the system | Researchers trained two state of the art RL agents, Rainbow DQN and A2C in a simulation to avoid lava. During training, the RL agent was able to avoid lava successfully and reach its goal. During testing, they slightly moved the position of the lava, but the RL agent was not able to avoid [24] |
| 4 | **Natural Adversarial Examples** | The system misrecognizes an input that was found using hard negative mining | Safety of the system | Here the authors show how by a simple process of hard negative mining[25], it is possible to confuse the ML system by relaying the example. |

---

| 5 | **Common Corruption** | The system is not able to handle common corruptions and perturbations such as tilting, zooming, or noisy images. | Safety of the system | The authors show how common corruptions such as changes to brightness, contrast, fog or noise added to images have a significant drop in metrics in image recognition.[26] |
| 6 | **Incomplete Testing in Realistic conditions** | The ML system is not tested in realistic conditions that it is meant to operate in. | Safety of the system | The authors highlight that that while defenders commonly account for robustness of the ML algorithm, they lose sight of realistic conditions. For instance, they argue that a missing stop sign knocked off in the wind (which is more realistic) than an attacker attempting to perturb the model.[25] |

## Acknowledgements

We would like to thank Andrew Marshall, Magnus Nystrom, John Walton, John Lambert, Sharon Xia, Andi Comissoneru, Jugal Parikh, Sharon Gillet, members of Microsoft's AI and Ethics in Engineering and Research (AETHER) committee's Security workstream, Amar Ashar, Samuel Klein, Jonathan Zittrain,  members of AI Safety Security Working Group at Berkman Klein for providing helpful feedback. We would also like to thank reviewers from 23 external partners, standards organization, and government organizations for shaping the taxonomy.

---

[26] Hendrycks, Dan, and Thomas Dietterich. "Benchmarking neural network robustness to common corruptions and perturbations." *arXiv preprint arXiv:1903.12261* (2019).